\documentclass[preprint,12pt]{elsarticle}

\usepackage{amssymb}
\usepackage{amsmath}
\usepackage{multirow}%
\usepackage{color}

\journal{xx}

\begin{document}

\begin{frontmatter}



\title{Deep Learning based Infrared Small Object
Segmentation: Challenges and Future Directions}


\author[1]{Zhengeng Yang\corref{cor1}} %
\ead{yzg050215@163.com}
\author[2]{Hongshan Yu\corref{cor1}} %
\ead{yuhongshancn@hotmail.com}
\cortext[cor1]{Corresponding authors}
\author[1]{Jianjun Zhang} 
\author[1]{Qiang Tang} 
\author[3]{Ajmal Mian} 



\affiliation[1]{organization={College of Engineering and Design,Hunan Normal University},
            addressline={}, 
            city={Changsha},
            postcode={410081}, 
            state={Hunan},
            country={China}}
  
\affiliation[2]{organization={School of Robotics,Hunan University},
            addressline={}, 
            city={Changsha},
            postcode={410082}, 
            state={Hunan},
            country={China}}            
\affiliation[3]{organization={Department of Computer Science,The University of Western Australia},
            addressline={}, 
            city={Perth},
            postcode={WA 6009}, 
            state={WA},
            country={Australia}}  
\begin{abstract}
Infrared sensing is a core method for supporting unmanned systems, such as autonomous vehicles and drones. Recently, infrared sensors have been widely deployed on mobile and stationary platforms for detection and classification of objects from long distances and in wide field of views. Given its success in the vision image analysis domain, deep learning has also been applied for object recognition in infrared images. However, techniques that have proven successful in visible light perception face new challenges in the infrared domain. These challenges
include extremely low signal-to-noise ratios in infrared images, very small and blurred objects of interest, and limited availability of labeled/unlabeled training data due to the specialized nature of infrared sensors. Numerous methods have been proposed in the literature for the detection and classification of small objects
in infrared images achieving varied levels of success. There is a need for a survey paper that critically analyzes existing
techniques in this domain, identifies unsolved challenges and provides future research directions. This paper fills the gap and offers a concise and insightful review of deep learning-based methods. It also identifies the challenges faced by existing
infrared object segmentation methods and provides a structured review of existing infrared perception methods from the perspective of these challenges and highlights the motivations behind the various approaches. Finally, this review suggests promising future directions based on recent advancements within this domain.

\end{abstract}

\begin{graphicalabstract}
\includegraphics{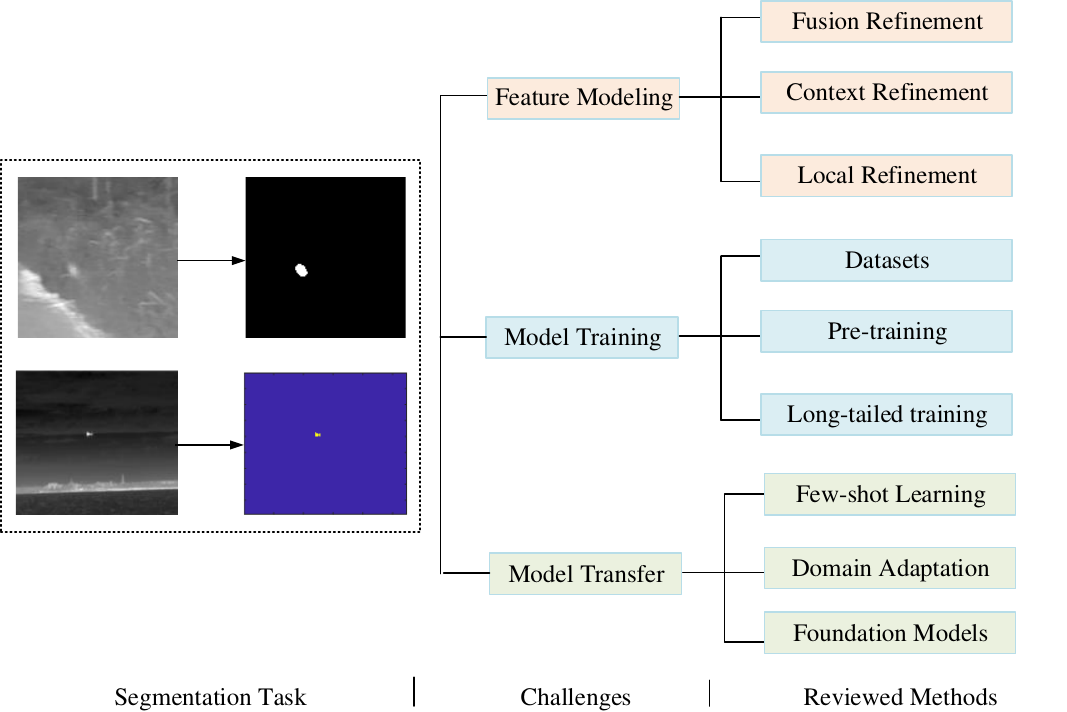}
\end{graphicalabstract}



\begin{keyword}


Infrared Perception\sep Dim Object Recognition\sep Semantic Segmentation\sep Deep Learning
\end{keyword}

\end{frontmatter}




\section{Introduction}
\label{seq_In}
Infrared imaging is widely used in fields such as autonomous driving, unmanned aerial vehicles (UAV), and surveillance due to its unique advantages in challenging scenarios such as low lighting, harsh weather, and long-distance sensing, making it an essential tool to support day and night sensing of various unmanned systems.  As a result, the application scenarios of infrared perception are becoming wider (ground-to-ground, ground-to-air, air-to-air, air-to-ground, air-to-sea) and the corresponding perception distances are becoming longer (such as long-distance rescue missions and illegal fishing vessel detection).

\begin{figure*}[!ht]
\centering
\includegraphics[width=5.5in]{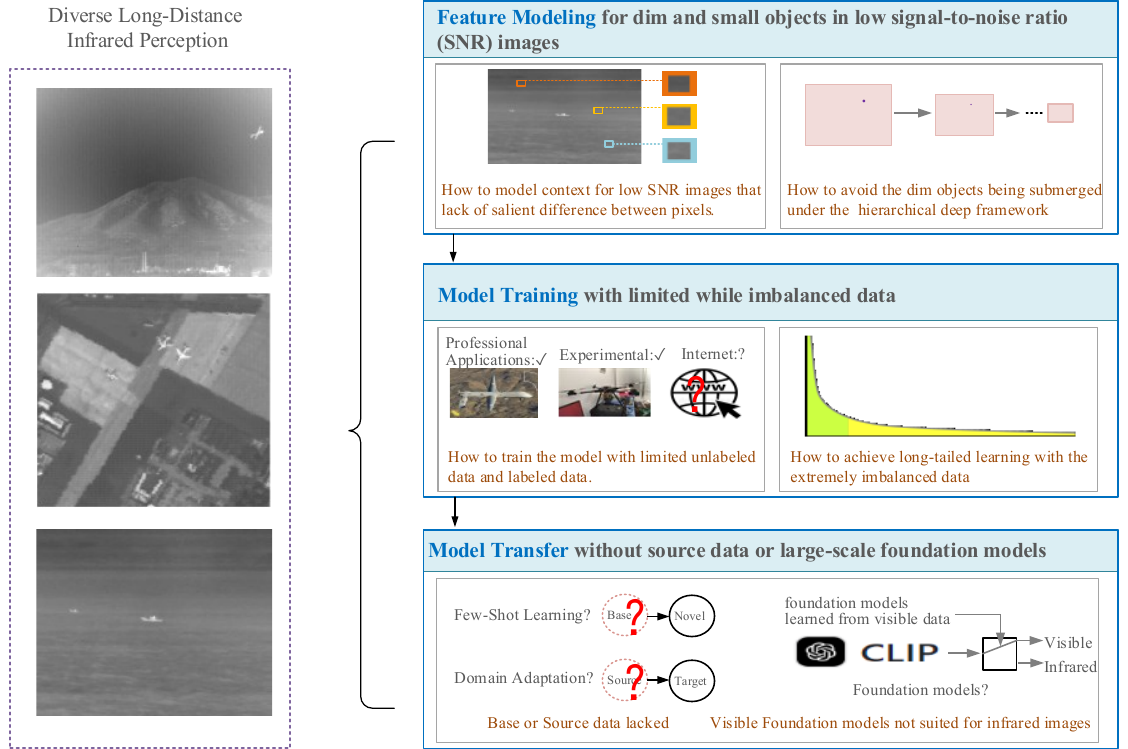}
\caption{The great challenges towards the goal of achieving high-quality infrared small object recognition of diverse long-distance perception.}
\label{fig_challenges}
\end{figure*}

In recent years, the rapid development of various multi-modal foundation models (e.g., CLIP~\cite{radford2021learning} and Align~\cite{jia2021scaling}) has shed light on achieving various visual perception tasks (e.g., semantic segmentation, object detection) with a unified model. However, the development of existing foundation visual perception models has benefited from the massive amount of visible light images with text descriptions on the Internet, making the distribution-sensitive deep foundation models not suited well for professional applications~\cite{ji2023segment}. On the other hand, for long-distance infrared perception scenarios represented by aerial object detection and ocean ship detection in Fig. \ref{fig_challenges}, the sensing images typically show a low signal-to-noise ratio, and the object of interest is usually extremely small and dim. 
In addition, due to the unique nature of long-distance infrared sensing applications, it is not easy to obtain a large-scale infrared dataset through the Internet.  
Therefore, confronted with these distinct challenges of infrared perception, achieving high-quality performance across diverse applications primarily involves overcoming three major challenges.

The first challenge is that of effective feature modeling for dim and small objects in low signal-to-noise ratio (SNR) images. Innovations in feature modeling, starting from SIFT~\cite{lowe1999object} to deep learning architectures like ResNet~\cite{he2016deep} and ViT~\cite{dosovitskiy2020image}, have driven significant advancements in visual perception. While the hierarchical abstraction in deep feature models enhances the high-level semantic representation of scene knowledge, it inevitably causes small objects to be gradually overshadowed by the background or larger objects. To address this, techniques like dilated convolution~\cite{chen2017deeplab,guo2024dmfnet}, which reduce feature subsampling, and multi-scale feature fusion~\cite{du2023bpr} have been employed to strengthen the representation of small objects. Despite these efforts, the recognition performance for dim and small objects remains significantly lower than for larger ones. 
Additionally, the success of feature pyramids~\cite{zhao2017pyramid} and Transformers~\cite{dosovitskiy2020image,liu2021swin} underscores the importance of semantic context in visual perception. However, long-distance infrared images with low SNR often contain few blurry objects of interest against a dim background, leading to a scarcity of contextual information. Therefore, the first grand challenge is to design a robust feature model that can generalize across diverse infrared perception scenarios and support fine-grained recognition of dim, small objects with limited contextual information.


The second challenge is that of model training with limited data. Deep learning models rely on large-scale datasets for effective training. However, long-distance infrared perception is often used in specialized applications like reconnaissance, requires expert annotations, making it difficult to gather sufficient labeled data for high-precision training. Recent advancements in unsupervised representation learning~\cite{he2020momentum,yang2022fully} have offered a promising solution by enabling model pre-training on large-scale unlabeled data to capture extensive common knowledge, thereby mitigating the challenge of limited annotations in practical applications. Unfortunately, acquiring large-scale unlabeled long-distance infrared data is also challenging due to the specialized nature of these applications. Consequently, obtaining such datasets also becomes difficult. Moreover, long-distance infrared images often feature very few dim, small objects, creating a significant imbalance between objects of interest and the background. Even with sufficient annotations, long-distance infrared perception faces the challenge of addressing long-tail recognition problems.


The third challenge is posed by model transfer across diverse application scenarios. Transferring an existing perception model to new but related scenarios or tasks is a well-established technique in machine learning. Recently, leveraging vast collections of visible light images paired with textual descriptions from the Internet, researchers have advanced multi-modal invariant representation learning through 'visual-text' alignment~\cite{radford2021learning}. This approach enables open-vocabulary recognition~\cite{wu2024towards}, where novel visual classes are defined by an unbounded (open) vocabulary and then aligned with unseen visual features, facilitating the recognition of previously unencountered classes.
However, machine learning is in general sensitive to data distributions and existing multi-modal foundation models trained on visible images are not yet suitable for specialized applications like long-distance infrared sensing. Additionally, the scarcity of multi-modal data, particularly textual descriptions of infrared images, poses a significant challenge to directly learning multi-modal invariant representations for long-distance infrared perception.

\begin{figure}[!ht]
\centering
\includegraphics[width=\textwidth]{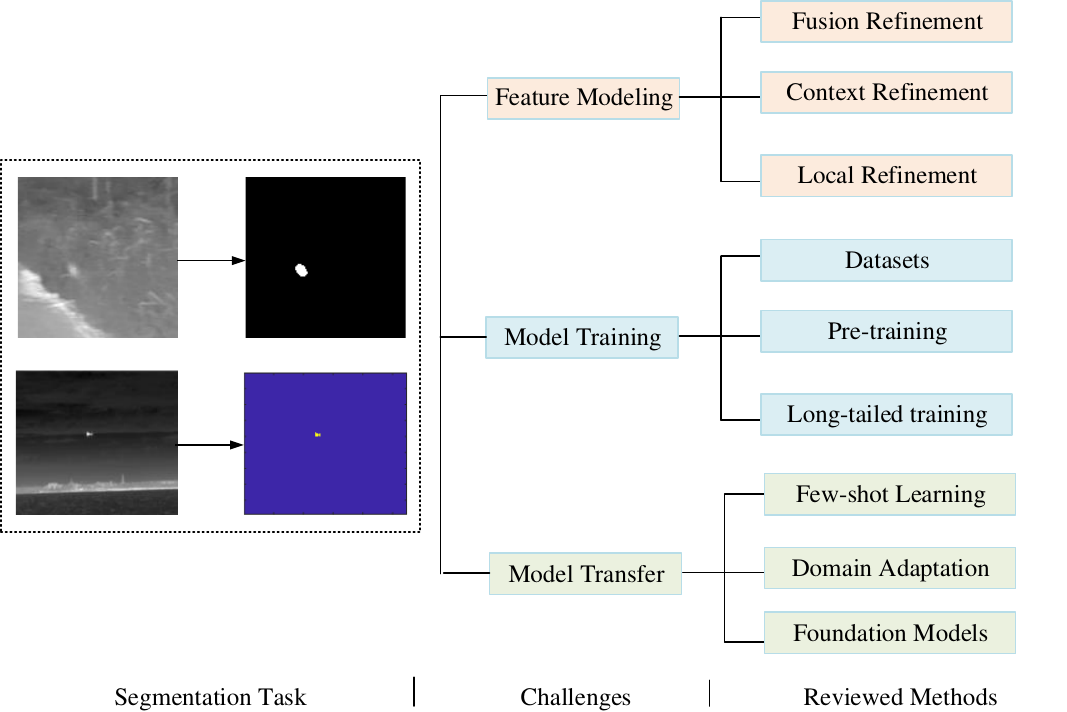}
\caption{The architecture of this review paper constructed from the perspective of challenges.}
\label{fig_archi}
\end{figure}

Given the above three grand challenges, this paper aims to provide a thorough review of recent efforts that contributed efforts to their solutions. Note that Kou et al.~\cite{kou2023infrared} provided the first survey on infrared small target segmentation networks. However, our survey differs from  Kou's review in two major aspects. First, we summarize the main challenges in achieving high-quality infrared small object recognition that can be extended to diverse long-distance infrared perception scenarios. We then systematically review existing methods that address or mitigate these challenges (Fig. \ref{fig_archi}). To our knowledge, this is the first structured review of infrared small object recognition methods from a challenge-based perspective, clarifying both the motivations behind these innovations and potential future directions. In addition, Kou's review mainly focused on methods that contributed to the first and second challenges while giving few analyses on the third challenge, which has received great attention recently.  Secondly, Kou's review directly replicated many existing model structures when introducing these methods, making it difficult for readers to discern their differences and similarities. In contrast, we have redrawn the representative model structures, focusing on highlighting their core innovations and differences from each other. Additionally, where possible, we have used a unified framework to redraw methods that belong to the same category. These contributions make the distinctions between the reviewed methods clearer and easier to understand.


\begin{figure}[!ht]
\centering
\includegraphics[width=3.1in]{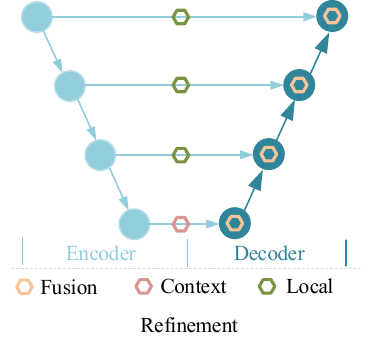}
\caption{The commonly used encoder-decoder structure as well as the refinement strategies for infrared small object recognition.}
\label{fig_ende}
\end{figure}

\section{Feature Modeling}
\label{sec_featuremodeling}
Feature modeling plays an essential role in the visual perception of unmanned systems. Well-known deep feature models represented by ResNet and ViT have greatly improved the visual perception accuracy of unmanned systems based on visible light images. Since both infrared perception and visible light perception can essentially be regarded as image processing, feature modeling methods for dim infrared object perception are mostly developed from existing visible light perception models. On the other hand, existing datasets for long-distance infrared perception mostly formulate the task as a binary segmentation problem of foreground dim objects. As a result, current feature models for long-distance infrared perception are mostly built on the ``Encoder-Decoder" framework (Fig. \ref{fig_ende}) ~\cite{dai2021asymmetric,Zhang2023dim2,zhang2022isnet,wang2023ccranet,zhang2023chfnet,zhang2023attention,yu2022pay,li2022dense,xi2024non,wang2024afe,chen2024feature,zhang2024single,guo2024dmfnet,zang2025dcanet,guo2024location,dai2024pick}  that is commonly used in visible light image segmentation~\cite{badrinarayanan2017segnet}.  Since the dim and small objects tend to overlap by background with the feature going deeper,  current methods mainly focus on how to retain as much information as possible for dim and small objects under the Encoder-Decoder framework. Efforts towards this goal can be roughly categorized into three classes we coined fusion of local details and high-level semantic features, semantic context refinement, and local details refinement.

\subsection{Preliminaries}
Since many methods have resorted to various attention mechanisms and feature pyramids to enhance the feature modeling on infrared dim small objects, we first introduce several commonly used attention models (e.g., channel attention, pixel attention, spatial attention, channel relation attention, and non-local attention) and feature pyramids that will be mentioned later. 
\subsubsection{Attention Models}
\begin{figure}[!h]
\centering
\includegraphics[width=\textwidth]{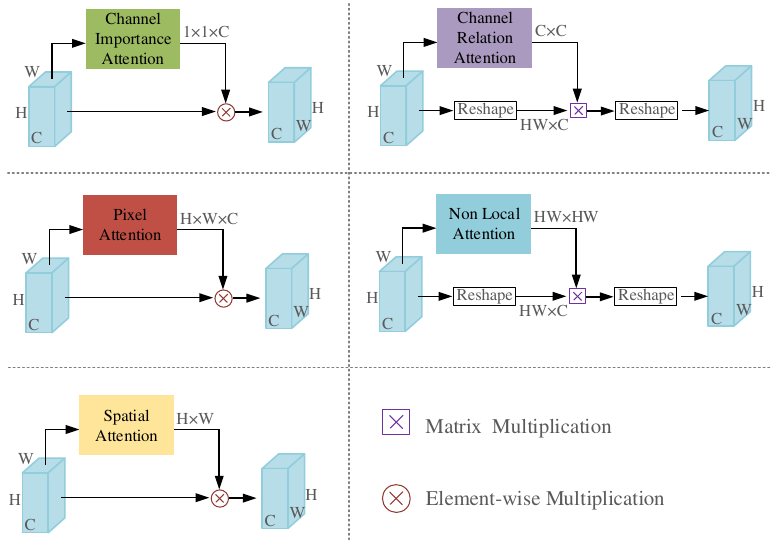}
\caption{Commonly used attention models in infrared perception.}
\label{fig_attention}
\end{figure}
Since the attention mechanism is usually used for modulating feature vectors, to simplify the difference between current attention mechanisms,  we clarified the core idea of their final modulation process in Fig.\ref{fig_attention}. Note that we do not give an analysis of the computation of modulation weights to make the difference easier to understand. 

Supposing $f_{h,w} \in R^C$ is a feature vector in a feature map $ F \in R ^{H\times W \times C}$ at spatial location $(h,w)$. 

The intuition of channel attention (CA)~\cite{hu2018squeeze} is that object recognition should pay more attention to some important channels. Thus, it performs feature modulation over $f_{h,w}$ by performing a Hadamard product, i.e., element-wise product, with a channel importance vector, and the importance vector is the same for all feature locations.

The pixel attention (PA)~\cite{zhao2020efficient} is similar to the channel attention, but the channel importance vector differs for different feature locations. 

The intuition of spatial attention (SA)~\cite{woo2018cbam} is that object recognition should pay more attention to some important spatial areas or pixels. Thus, it performs the feature modulation over $f_{h,w}$ by multiplying it with a spatial importance weight scalar. 

The intuition of non-local attention (NLA), also named self-attention~\cite{wang2018non} is that a feature location should pay attention to all feature locations that may influence its representation (e.g., a feature of a boat can be enhanced by its surrounding water features). Thus, it performs feature modulation over $f_{h,w}$ by aggregating features from all $HW$ feature vectors after multiplying each of them with a weight scalar.  

The channel-wise cross attention (CCA)~\cite{wang2022uctransnet} can be easily explained if we name non-local attention as pixel-wise cross attention. In other words, the non-local attention performs feature modulation over $f_{h,w}$ by considering pixel-wise cross relations, while the channel-wise cross attention performs feature modulation by considering channel-wise relations.

\subsubsection{Feature Pyramids}

\begin{figure}[!h]
\centering
\includegraphics[width=\textwidth]{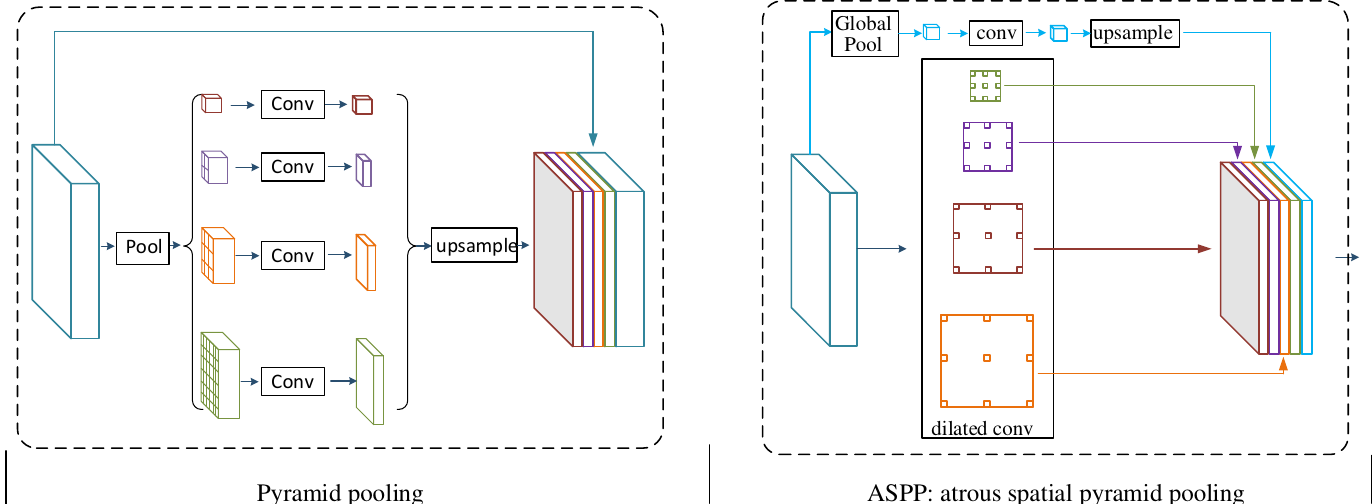}
\caption{Commonly used feature pyramids in deep learning.}
\label{fig_pyramid}
\end{figure}

The commonly used feature pyramids include the spatial pyramid pooling and atrous spatial pyramid pooling shown in Fig. \ref{fig_pyramid}. The spatial pyramid pooling for the segmentation task~\cite{zhao2017pyramid} pools the feature to different resolutions and then employs feature transform to them independently. Finally, these multi-scale features are concatenated together after resolution alignment. To improve the efficiency of feature pyramids, the ASPP module employs dilated convolutions with different dilation rates to obtain multi-scale receptive fields.

\subsection{Fusion of Local Details and High-level Semantic Features}
\label{feature_fusion}
\begin{figure*}[!h]
\centering
\includegraphics[width=\textwidth]{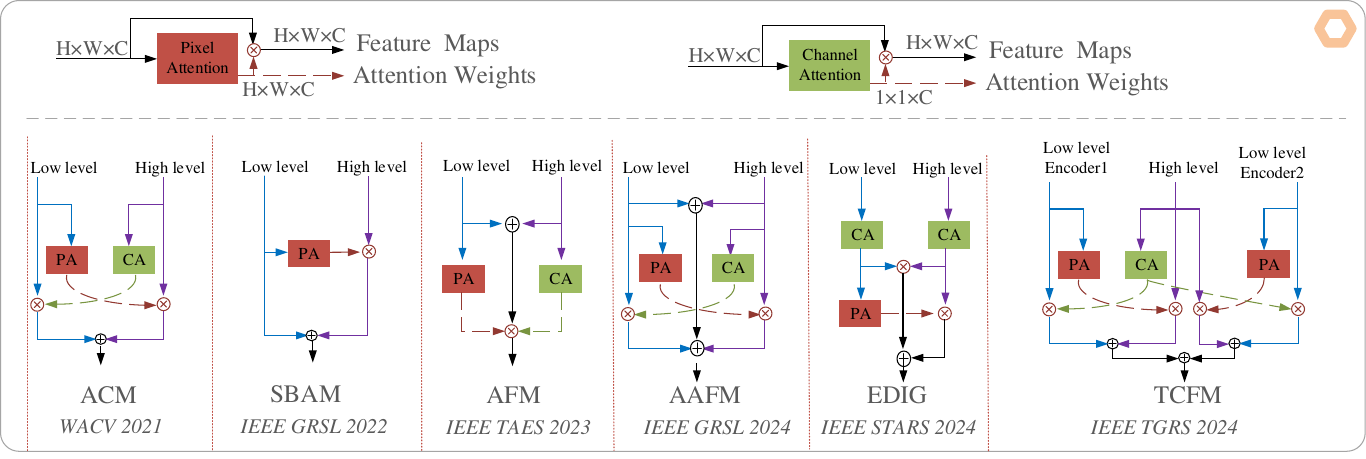}
\caption{Representative fusion refinement methods, including Simplified Bilinear Interpolation Attention Module (SBAM)~\cite{yu2022pay}, Asymmetric Contextual Modulation (ACM)~\cite{dai2021asymmetric}, Asymmetric Fusion Module (AFM)~\cite{zhang2023attention}, Asymmetric Attention Fusion Module (AAFM)~\cite{xi2024non}, Encoder
and Decoder Interactive Guidance (EDIG)~\cite{wang2024afe}, and Triple Cross-Layer Fusion Module (TCFM)~\cite{guo2024dmfnet}. Note that the attention mechanisms were usually used for feature modulation and can be used either for self-modulation or cross-modulation, we differentiate these two modulations with solid and virtual lines, respectively. In particular, if the attention model is followed by a solid line, it means the attention is used to perform a self-modulation over the input feature, otherwise, it means the attention model outputs attention weights for cross-level modulation. }
\label{fig_furefine}
\end{figure*}

As shown in Fig. \ref{fig_ende} the basic Encoder-Decoder structure uses skip connections to transfer spatial details to high-level semantic features, but traditional addition or concatenation-based feature fusion would lead a sub-optimal results due to the possible conflict and redundancy between fused features. To this end, Yu et al. proposed a simple bilinear interpolation attention fusion module named SBAM~\cite{yu2022pay}. This method used spatial detail features to calculate pixel attention and then employed it to modulate high-level semantic features, which essentially led to the effect of forcing the high-level feature fusion to pay much more attention to spatial areas of dim small objects. Inspired by the
sensitivity of the human vision system to contrast differences, Chen et al.~\cite{chen2024feature} proposed an Orthogonal Central Difference Fusion (OCDF) module  to refine the high-level semantic features with two contrast difference attention maps generated from low-level features along orthogonal directions, i.e., row or column
direction.

However, the computation of attention relies on high-quality spatial details. To this end, Dai. et al presented an asymmetric context fusion module named ACM~\cite{dai2021asymmetric},  which adopted two asymmetric fusion branches to achieve cross-modulation between spatial details and high-level semantics. In particular, the top-to-down fusion tries to improve the spatial details by fusing high-level semantic contexts through channel attention, while the down-to-top fusion aims to fuse more salient spatial details obtained by pixel attention into the high-level semantic features. A recent work named PGA (Patch-group Attention)~\cite{zhang2024single} replaced channel attention and pixel attention in ACM with group channel attention and patch spatial attention respectively.  Zhang et al.~\cite{zhang2023attention} argued that the two asymmetric fusion branches in ACM are independent to each other, which would lead to the high-level semantic features lacking the modeling of channel importance while the spatial details lack the modeling of object saliency. To this end, they proposed applying channel attention and pixel attention to high-level semantic features and spatial details, respectively and then employed the asymmetric fusion strategy.



For the sake of clarity, we do not introduce more methods here and instead list representative methods as well as their ideas in this direction in Table \ref{table_furefinemethods}.

\begin{table}[!h]
\renewcommand\arraystretch{1.5}
\scriptsize
\caption{Representative fusion refinement methods. HLF and LLF mean high-level features and low-level features respectively. SA, CA, and PA mean spatial attention, channel attention, and pixel attention, respectively.}
\centering
\hspace*{-3.5cm}
\label{table_furefinemethods}
\begin{tabular}{ll}
\hline
Methods &Novelties different from others \\
\hline
Baseline& Direct concatenation or element-wise summation (baseline fusion for short)\\
FAMCA~\cite{huang2021infrared}& Perform CA \textbf{after} the baseline fusion.\\
SBAM~\cite{yu2022pay}&  Perform modulation over HLF with PA weights computed from LLF \textbf{before} the baseline fusion. \\
BLAM~\cite{dai2021attentional}& Perform modulation over HLF with PA weights computed from LLF \textbf{before} the baseline fusion. \\
TOAA~\cite{zhang2022isnet}& Similar to the SBAM, but the modulation is performed twice by computing attention along the row and column direction of LLF\\
MCAF~\cite{nian2023local}& Perform CA over LLF first and then modulate LLF with PA weights computed from HLF  \\
ACM~\cite{dai2021asymmetric}& Perform cross-modulation between LLF and HLF with CA and PA weights 
\textbf{before} the baseline fusion                     \\
PGA~\cite{zhang2024single}&    Perform cross-modulation between LLF and HLF with group-wise CA and patch-wise SA weights \textbf{before} the baseline fusion                 \\
AFM ~\cite{zhang2023attention} &  Perform modulation over baseline fusion with PA and CA weights computed respectively from LLF and HLF\\

MFFM~\cite{wang2023multi} &  Perform cascaded feature fusion from high-level semantics to low-level details \\
AAF~\cite{zhang2024global}& Similar to the AFM\\
AAFM~\cite{xi2024non}& The combination of ACM and AFM\\
EDIG~\cite{wang2024afe}& Perform CA over LLF and HLF, respectively, and multiply them. Then, fused with HLF modulated by PA weights computed from LLF.  \\
TCFM~\cite{guo2024dmfnet}& Employed dual encoder, and then performed ACM twice for encoder-decoder fusion \\


\hline
\end{tabular}

\end{table}

\subsection{Semantic Context Refinement}
\begin{figure}[!h]
\centering
\hspace*{-1.5cm}
\includegraphics[width=6 in]{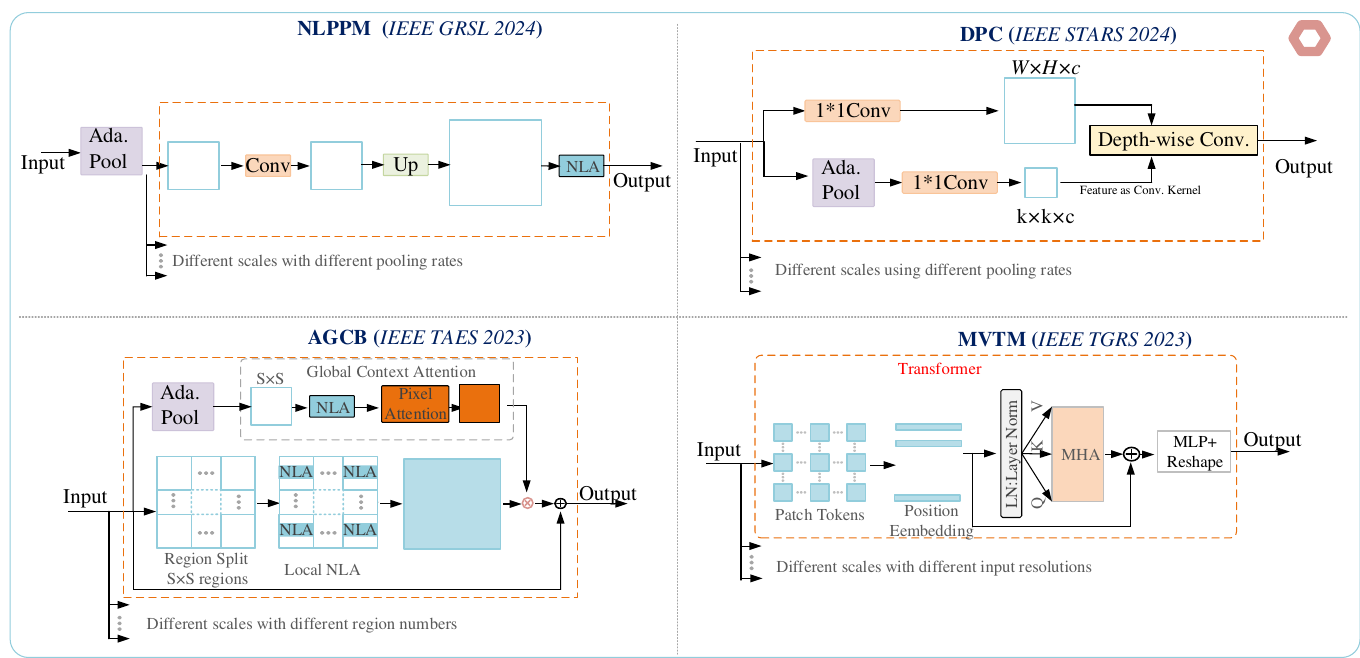}
\caption{Representative context refinement methods, including Non-Local Enhanced Pyramid Pooling Module (NLPPM)~\cite{xi2024non}, Dynamic Pyramid Context (DPC)~\cite{chen2024dynamic},Attention-Guided Context Block (AGCB)~\cite{zhang2023attention}, and Multilevel ViT Module (MVTM)~\cite{wu2023mtu}. The NLA means non-local attention, and the MHA means multi-head attention. Note that we have simplified the drawings presented in the original paper by considering only the core idea here, and more details can be referred to the corresponding papers. }
\label{fig_conrefine}
\end{figure}

It is well-known that dependency modeling between local information, i.e., context modeling, plays a key role in feature modeling. Inspired by the feature pyramid commonly used in visible light perception, Xi et al. devised a  Non-Local Enhanced Pyramid Pooling Module (NLPPM)~\cite{xi2024non} for infrared small object segmentation.  The NLPPM is constructed by adding a non-local attention (NLA) operation to each scale of the standard pyramid pooling (left of Fig. \ref{fig_pyramid}).  Similarly, Zhang et al.~\cite{zhang2023attention} presented an attention-guided context pyramid structure named AGCB. Different from the NLPPM, which performs NLA at the scale level, the AGCB performs NLA at the region level (see Fig. \ref{fig_conrefine}). A similar idea was also introduced in LSPM~\cite{huang2021infrared}. Although the multi-scale local self-attention improves the local context modeling for dim and small objects, it suffers from a high computational cost. Recently, Chen et al. designed an efficient pyramid named dynamic pyramid context (DPC)~\cite{chen2024dynamic}. As shown in Fig.\ref{fig_conrefine}, for each pyramid scale, the DPC module directly generates depth-wise convolution kernels ($k\times k \times c$) from the scaled features ( $k\times k \times c$), which was then used to capture information from the un-scaled features transformed from the input.    


\begin{table}
\renewcommand\arraystretch{1.5}
\caption{Representative context refinement methods. ASPP, FPN, PA, NLA, CA, and SA mean atrous spatial pyramid, feature pyramid network, pixel attention, non-local attention, channel attention, and spatial attention, respectively.}
\scriptsize

\centering
\hspace*{-3.5cm}
\begin{tabular}{ll}
\hline
Methods &Novelties different from others \\
\hline
Classical & Employ feature pyramid models such as ASPP and FPN\\
AGCB~\cite{zhang2023attention}& Enhancing each scale before pyramid construction by performing global PA and local NLA\\

NLPPM~\cite{xi2024non}& Enhancing each scale before pyramid construction by performing NLA\\
LSPM~\cite{huang2021infrared}& Enhancing each scale before pyramid construction by performing pixel-to-region attention (a variant of NLA) based modulation\\
MVTM& Enhancing each scale before pyramid construction by performing multi-head attention, i.e., transformer, over visual patch tokens \\
CFAR~\cite{wang2024afe}&  Perform NLA and a cross-layer NLA sequentially \\
CBAM~\cite{zhang2024infrared}& Perform CA and SA sequentially.\\
REAM ~\cite{guo2024dmfnet}& Perform CA after ASPP\\
DPC ~\cite{chen2024dynamic}& Employing depth-wise convolution with filters generated directly from scaled features \\
\hline
\end{tabular}
\label{table_conrefinemethods}
\end{table}

In recent years, with the success of visual Transformers, employing Transformer to model context dependency for infrared images has received great attention. Since the visual Transformer relies on salient information to model context between visual tokens, existing transformer-based long-distance infrared image recognition methods typically extract features through CNN, and then apply the Transformer to capture the context dependence between feature tokens. For example, Liu et al.~\cite{liu2023infrared} were the first to apply the Transformer to model the long-distance dependency of convolutional features of infrared dim and small objects.  Subsequently, Wu et al.~\cite{wu2023mtu} introduced the idea of multi-scale fusion and constructed a feature pyramid coined MVTM  using the well-known visual transformer model ViT. The MVTM divided the CNN features into blocks of different scales, and then used the ViT structure to model the dependencies between these multi-scale feature blocks.

It can be observed from Table \ref{table_conrefinemethods} that many context refinement strategies usually resort to pyramid feature fusion. Thus, these methods can also be classified into the first class, i.e., the fusion of spatial details and high-level semantic information from a broad perspective. However, the level of spatial details considered in these methods is relatively higher for the goal of semantic context modeling.

\subsection{Local Details Refinement}

\begin{figure*}[!h]
\centering
\includegraphics[width=\textwidth]{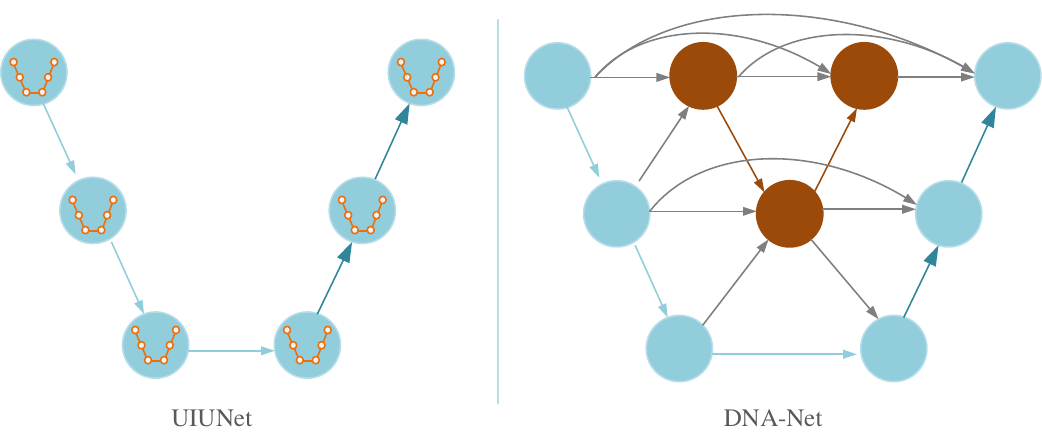}
\caption{Structures of UIUNet and DNA-Net, two well-known methods that can preserve local details by employing sub-UNets to achieve flexible local receptive fields. Note that we simplified the structure of UIUNet~\cite{wu2022uiu} and DNA-Net~\cite{li2022dense} by omitting two scales.  }
\label{fig_derefine2}
\end{figure*}

The above fusion strategies and semantic context modeling both rely on accurate spatial details. To this end, Bruce et al.~\cite{mcintosh2020infrared} defined a metric to evaluate the object-to-clutter ratio and proposed to optimize the parameters of the first convolution layer by maximizing this metric, which can enhance the dim and small object representation while ensuring the minimum clutter energy. This method improved the feature modeling for dim and small objects by considering more characteristics of infrared images. However, integrating traditional feature modeling means this method cannot be trained end-to-end. To enhance the local details extraction of the end-to-end encoder-decoder structure,  Wu et al.~\cite{wu2022uiu} embedded a smaller-scale encoder-decoder structure into the encoder to achieve high-precision extraction of local information, which formed the well-known UIUNet for dim and small object recognition (left of Fig. \ref{fig_derefine2}). Instead of embedding sub-UNet in UNet, Li et al.~\cite{li2022dense} stacked multiple sub-UNet and built a dense nested structure named DNA-Net. These sub-UNets with different depths are useful to capture details of objects of different sizes.
Chen et al.~\cite{chen2024feature} developed a dual branch encoder structure, which consists of a lightweight high-resolution branch and a lower-resolution branch, to better preserve details of small objects. Considering the increasing complexity of the encoder would pose challenges for deployment, Zhang et al.~\cite{zhang2024mrf} devised a multiple perception encoder (MPE) that performs parallel convolutions over grouped features with different kernel sizes. Similarly, Dai et al. devised a dilated difference convolution (DDC)~\cite{dai2024pick} that consists of three parallel convolutions of specialized goals (e.g., edge preservation). Guo et al. ~\cite{guo2024fcnet} integrated dilated convolutions and deformable convolutions~\cite{dai2017deformable} to achieve flexible local receptive fields. 

\begin{table}
\renewcommand\arraystretch{1.5}
\caption{Representative local refinement methods. ASPP, FPN, PA, NLA, CA, CCA, and SA mean atrous spatial pyramid, feature pyramid network, pixel attention, non-local attention, channel attention, channel-wise cross attention, and spatial attention, respectively.}
\scriptsize
\centering
\hspace*{-3.5cm}
\begin{tabular}{ll}
\hline
Methods &Novelties different from others \\
\hline
OTC & Employ an independent learned layer through an object-to-ratio based optimization\\
ICA~\cite{zhang2024infrared}& Perform row-level and column-level CA. respectively.\\
Embedded UNet~\cite{wu2022uiu}  & Embedding small scale UNet into the feature layers\\
MALC~\cite{nian2023local}& Employ a local ASPP and then perform PA\\
MLFPM~\cite{wang2023multi}& Employ feature pyramid at each scale\\
FEM~\cite{guo2024fcnet}& Employ dilated convolutions and deformable convolutions to achieve flexible local receptive fields\\
SAE~\cite{quan2024hmnet}& Employ CA and SA sequentially\\
SiAM/PiAM~\cite{zhong2024hierarchical} & Perform feature transformation by NLA and CCA\\
Swin Transformer+GCL~\cite{zhu2024towards} & Employ Swin Transformer~\cite{liu2021swin} as the feature network and an edge-prior guided feature refinement strategy \\
Swin Transformer+LCA~\cite{zhao2023res} & Employ Swin Transformer~\cite{liu2021swin} as the feature network and a local contrast attention module\\
MDAF~\cite{li2024moderately} & Employ larger convolutional kernels and dense connections in the basic residual block of ResNet.\\
MRCB~\cite{li2024cross} & Employ dilated convolution and CA in the basic residual block of ResNet. \\
Contrast-shape Encoder~\cite{lin2024learning}& Employ central difference convolution~\cite{yu2020searching} and large kernel convolution\\
SCTB ~\cite{yuan2024sctransnet}& Perform CCA between single-level feature and concatenated multi-scale features  \\
\hline
\end{tabular}
\label{table_derefinemethods}
\end{table}

\begin{figure*}[!h]
\centering
\includegraphics[width=\textwidth]{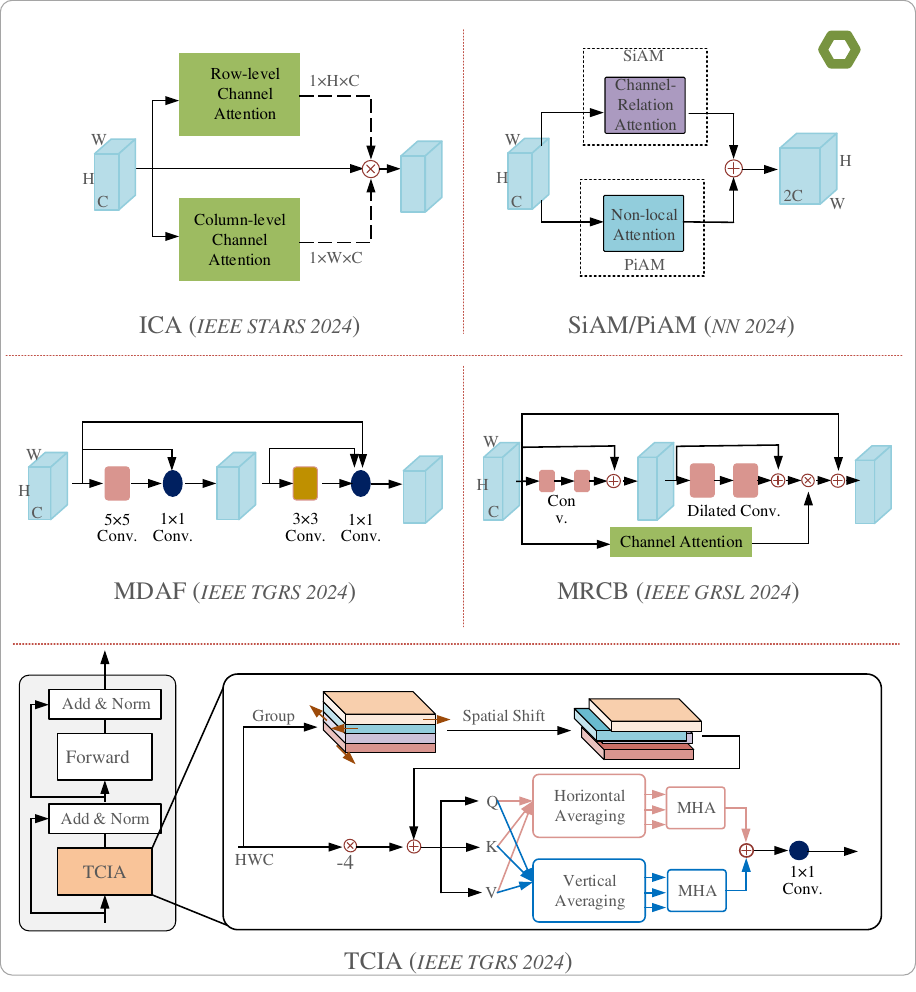}
\caption{Representative local refinement methods, including ICA(improved coordinate attention)~\cite{zhang2024infrared}, SiAM/PiAM(spatial/pixel-interactive attention module)~\cite{zhong2024hierarchical},MDAF(moderately dense adaptive feature fusion)~\cite{li2024moderately}, MRCB(multiscale residual connection block)~\cite{li2024cross}, TCIA(thermal conduction-inspired attention)~\cite{chen2024tci}. Similar to Fig .\ref{fig_conrefine}, we have simplified the drawings presented in the original paper by considering only the core idea here. MHA means multi-head attention used in Transformer}
\label{fig_derefine}
\end{figure*}

It is not difficult to find that methods mentioned above, including stacked UNet in DNA-Net~\cite{li2022dense}, dual branch with different feature resolutions~\cite{chen2024feature}, MPE~\cite{zhang2024mrf}, and deformable convolution~\cite{guo2024fcnet}, mostly aims to achieve flexible local receptive fields to adaptively preserve details of objects of different sizes. Besides these methods, similar to fusion refinement and semantics refinement, the attention mechanism is also widely used to enhance local features directly. For example, Zhong et al.~\cite{zhong2024hierarchical} presented two attention modules named spatial interactive attention and pixel interactive attention to improve local feature modeling (top right of Fig. \ref{fig_derefine}). In particular, the spatial interactive attention aims to enhance the spatial information interaction between feature channels, and the pixel interactive attention aims to improve local context modeling via a non-local attention operation. Similarly, the Cross Aggregation Encoder (CAE) proposed in \cite{chen2024local} also employed two branches to enhance the local saliency and context modeling simultaneously. The local saliency is enhanced by calculating the local contrast that can be formulated by a convolution operation with fixed weights, and the context modeling is enhanced by employing multi-head self-attention to local features. Considering that frequency clues are
more sensitive to minor gradient differences exhibited in infrared images, Chen et al. ~\cite{chen2024freqodes} employed interaction between spatial domain and frequency domain to enhance local saliency and context modeling, which is achieved by modulating features of each domain with self-attention computed from the other domain.   Zhao et al.~\cite{zhao2023res} developed a local contrast attention (LCA) module to enhance directly the local feature extraction of the well-known swim transformer. The core idea is to subtract the original features with its features shifted from two converse directions to get local contrast feature maps. Recently, Chen et al. ~\cite{chen2024tci} proposed a novel perspective to mimic the feature evolution in deep learning by a thermal conduction process, which yields a feature extraction rule that pixel features in a certain layer are propagated (similar to the heat conduction) from its surrounding pixels in the former layer along either x-axis or y-axis. Such propagation is then approximated by a thermal conduction-inspired attention (TCIA) module (bottom of Fig. \ref{fig_derefine}) that first employs a spatial shift operation to produce surrounding feature groups and then performs directional feature aggregation with horizontal and vertical self-attention.


Similarly, we listed the difference of existing local details refinement methods in Table \ref{table_derefinemethods}.

\subsection{Other methods}
\begin{figure*}[!h]
\centering
\includegraphics[width=\textwidth]{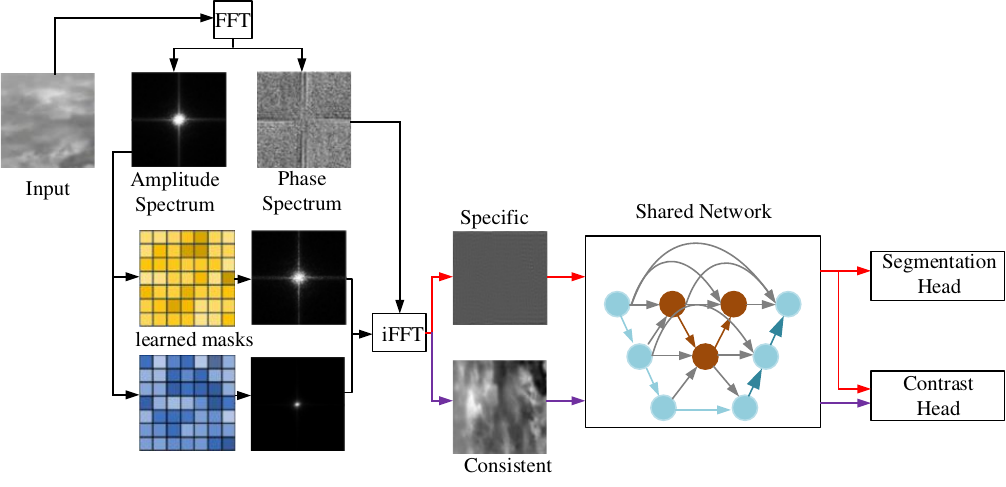}
\caption{The simplified architecture of FDDBANet~\cite{Huang2024FDDBA}, which proposed a novel idea to decouple the feature extraction of specific objects from backgrounds with frequency domain decoupling}
\label{fig_fddba}
\end{figure*}

This subsection will introduce some interesting ideas that were not mentioned in the above. These ideas typically resort to some traditional methods to facilitate feature learning over small objects.  
For example, Huang et al.~\cite{Huang2024FDDBA} proposed a novel idea to decouple the feature extraction of objects from backgrounds, which produces two input images containing either only specific objects or only backgrounds, and then maximize the feature disparity between objects and backgrounds through contrast learning during the segmentation process. As shown in Fig. \ref{fig_fddba}, the decoupled object input and background input are obtained by employing two learned masks to the spectral components decomposed from infrared input with Fast Fourier Transform (FFT). Tang et al.~\cite{tang2024LDCNet} proposed a similar decomposition idea from the perspective of low-rank representation. In particular, they employ the robust principal component analysis (RPCA) to separate the low-rank background matrix from the infrared image. Then, the difference matrix between the background and infrared input is used to learn a series of differential attention matrices to guide the feature learning on infrared images. Instead of using the RPCA for input decomposition, Wu et al.~\cite{Wu_2024_WACV} treat the infrared small object segmentation as an iterative RPCA problem and then unfold the iterative process via deep networks.

Most methods mentioned so far mainly focused on improving the accuracy of feature modeling and paid less attention to the efficiency of infrared small object segmentation. Recently, Zhang et al. \cite{zhang2024irprunedet} presented a wavelet channel pruning strategy to the basic UNet architecture and achieved a trade-off between accuracy and efficiency. Considering that many methods have used computationally expensive visual transformers to improve global or local context modeling, Chen et al. \cite{chen2024mim} first applied Mamba~\cite{gu2023mamba}, an efficient and popular alternative to the transformer, to infrared small object segmentation for context modeling. To tailor the Mamba for small object recognition, the authors added an inner Mamba to model dependency between sub-patches inside each image patch. Together with the outer Mamba targeted for dependency modeling between patches, the formed Mamba-in-Mamba achieved the goal of guaranteeing
higher efficiency while sufficiently extracting both local
and global information.

\subsection{Summary and the Future}
In summary, existing efforts contributed to infrared dim and small object feature modeling mainly focused on exploring better strategies for the extraction and fusion of spatial details and high-level context. Although these methods have improved the recognition accuracy of dim and small targets, their ideas are typically co-opted from visible light image perception, without fully considering the characteristics of long-distance infrared images. In addition, the above various strategies usually need to be used jointly (will be detailed later in Table \ref{table_allmethods}) to ensure the recognition ability of dim and small targets, resulting in a complex model with high computational cost. 
Therefore, exploiting the unique characteristics of infrared images that are different from visible light images for efficient feature modeling would be a promising future direction. 

\section{Model Training}
\label{sec_modeltraining}
It is well-known that the training of parametric deep models relies on a large amount of data. As mentioned in the Introduction section, the long-distance infrared perception lacks both unlabeled data and labeled data. This section will mainly focus on methods or strategies that aim at solving the training issues of infrared dim and small object recognition, including existing datasets, pre-training strategies, and long-tailed learning methods.

\begin{table}[ht]
\renewcommand\arraystretch{1.5}
\caption{ Existing datasets of infrared small object recognition.}

\centering

\begin{tabular}{l  l l  l l}
\hline
\multirow{2}{*}{Datasets}  & \multicolumn{3}{c}{Number of Samples} &\multirow{2}{*}{Types of Samples}\\
\cline{2-4}
 &Total &Training &Test\\
\hline
MDvsFA~\cite{wang2019miss} &10100 &10000 & 100 &Single Image\\
NUAA-SIRST~\cite{dai2021asymmetric} &427 &341 &83 &Single Image\\\
NCHU-SIRST~\cite{shi2023infrared} &500 &273 &317 &Single Image\\\
IRSTD-1k~\cite{zhang2022isnet} &1000 &800 &201 &Single Image\\
NUDT-SIRST~\cite{li2022dense} &1326 &1061 &265 &Single Image\\
NUDT-SIRST-sea~\cite{wu2023mtu} &48 & & &Single Image\\

Anti-UAV~\cite{jiang2021anti} &318 & & &Sequences\\
IST-A~\cite{xu2023multi}  &158 & & & Sequences\\
\hline

\end{tabular}
\label{table_dataset}
\end{table}

\subsection{Datasets}
Although data is the cornerstone of deep learning, large-scale datasets for long-distance infrared perception are still relatively lacking due to the application specificity. Wang et al.~\cite{wang2019miss} open-sourced the first dataset (MDvsFA) for infrared dim and small target perception. This dataset contains nearly 10,000 training images and 100 test images. Note that a large number of its images are synthetic images that have obvious illogical annotations. In addition,  the synthetic small objects produced from Gaussian distribution tend to make the object labeling range inflated erroneously. As a result, although the synthetic dataset provides a significantly larger training set than real dataset such as NUAA-SIRST will be mentioned later, the performances achieved on MDvsFA are typically significantly lower than those achieved on real dataset. For instance, the UIU-Net achieved a 78.25\% IoU on the NUAA-SIRST dataset while achieved only 47.73\% IoU on the MDvsFA dataset with the same framework. To produce high-quality synthesized images, Li et al. incorporated an adaptive target size function and an adaptive intensity and blur function to avoid unreasonable and unrealistic images, respectively, which formed another synthesized dataset coined NUDT-SIRST. The NUDT-SIRST dataset~\cite{li2022dense} contains 1,326 (including 1061 for training and 265 for testing) infrared images of different scenes. Performance achieved on NUDT-SIRST typically close to those achieved on real dataset NUAA-SIRST, indicating that synthesized images in NUDT-SIRST can be viewed as samples taking from the same distribution.  

Dai et al.~\cite{dai2021asymmetric} developed the first dataset (NUAA-SIRST) consisting of real long-distance infrared images. However, the dataset only contains 427 pictures and 480 infrared dim and small object instances. Shi et al. contributed a similar small-scale dataset called NCHU-SIRST~\cite{shi2023infrared}, which contains only 500 real images. Subsequently, IRSTD-1k~\cite{zhang2022isnet} expanded the real infrared image samples to 1,000 by considering more scenes (oceans, rivers, mountains, cities, clouds) and more classes (drone, animals, vehicles, etc.)   The same team of NUDT-SIRST collected a space-based dim and small ship detection dataset named NUDT-SIRST-Sea~\cite{wu2023mtu}, containing only 48 spatial maritime infrared images. 

The above-mentioned datasets are usually targeted for single-frame infrared perception and thus the samples are usually discontinuous in sensing time. In light of that the long-distance infrared perception usually needs to track the interested dim and small objects, there are also some infrared datasets consisting of sequences or videos.  For example, Hui et al.~\cite{hui2020dataset} developed an infrared dataset for dim and small aircraft detection and tracking. The dataset consists of 22 video sequences, containing 16,177 frames of images and 16,944 aircraft instances. Similarly, Jiang et al.~\cite{jiang2021anti} developed a dataset consisting of 318 videos for infrared drone detection. Both these two datasets consider a unitary scene, in the contrary, the IST-A dataset~\cite{xu2023multi} collected 158 sequences of multiple infrared perception scenes. 

So far, the largest dataset for infrared small object recognition is the IRDST~\cite{sun2023receptive} built by Sun et al. The IRDST contains 142,727 frames comprising 40,650 real frames and 102,077 simulated frames respectively.

\subsection{Pre-training}
Applying pre-training on large-scale datasets to learn common knowledge that can be transferred to various tasks has been common sense in visual perception, especially after the unsupervised feature learning making pre-training on large-scale unlabeled data possible. However, due to the lack of large-scale unlabeled and labeled datasets, existing methods for infrared dim and small object recognition usually do not conduct pre-training. A recent work~\cite{lu2024sirst} proposed to apply Noise2Noise displacement to real long-distance infrared images to produce a large-scale synthetic dataset for self-supervised training. Besides this work, most methods directly adopted the pre-trained visible light features model or trained the feature and recognition network in an end-to-end manner. 

In 2023, the Facebook AI Research (FAIR) group proposed the first large-scale pre-training model (ImageBind)~\cite{girdhar2023imagebind} that can be used for infrared image perception. Based on the well-known CLIP model, the ImageBind achieved the alignment of six modalities including text, visible light, sound, depth, infrared thermal and IMU. The core idea of this method is to align all the modalities that CLIP has not learned with the visual modality of CLIP. Thus, for the alignment of two modality, it requires a multi-modal dataset that contains large-scale paired samples of these two modalities. For the interested infrared modality of this review, ImageBind adopted the LLVIP dataset, which contains 16,836 ``infrared-visible light" pairs, to learn to align the infrared thermal modality with visible light modality. Note that the infrared images in the LLVIP~\cite{jia2021llvip} dataset are close-range thermal infrared images that has a data distribution significantly different from the long-distance dim infrared images, thus it is difficult to directly apply this foundation model to infrared dim and small object recognition.

In 2024, Zhang et al.~\cite{zhang2024irsam} proposed to apply the well-known pre-trained segment anything (SAM) model~\cite{kirillov2023segment} for infrared small object recognition. Considering that it is difficult to obtain prompts for infrared small object recognition, the authors directly enhanced
SAM’s encoder-decoder architecture with feature modules specialized for small infrared objects and then fine-tuned them on infrared datasets.

\subsection{Long-tailed Training}
Long-tail distribution means that the number of samples of some classes only accounts for a tiny proportion of the total number of samples. If we draw a curve with the proportion number of each class from large to small, the curve at tail classes will presented in a ``long" straight line due to the small change in the proportion. Since the objects in the long-distance infrared images are dim and small, the distribution of interested objects and the background is extremely unbalanced, resulting in the long-tailed problem. To achieve effective training under long-tailed distribution, current methods either employ data augmentation or weighted loss function as follows.

\begin{figure*}[!h]
\centering
\includegraphics[width=\textwidth]{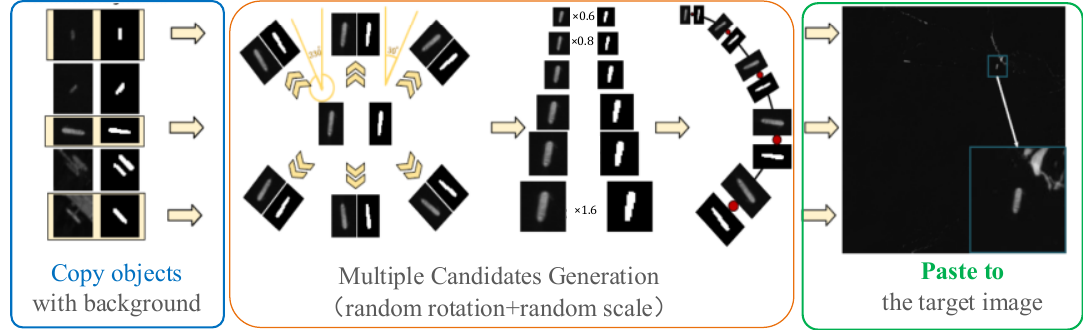}
\caption{The copy-and-paste data augmentation used for infrared ship detection.}
\label{fig_copypaste}
\end{figure*}
\subsubsection{Data Augmentation} The most straightforward way to solve the long-tailed problem is to oversample tail classes during training, which is usually achieved by data augmentation in the deep learning era. One of the most well-known data augmentation strategies for deep learning is the copy-and-paste ~\cite{ghiasi2021simple}. The core idea of this strategy is to first copy object instances from arbitrary training images and then paste these instances into arbitrary training images. Wu et al.~\cite{wu2023mtu} were among the first to introduce the copy-and-paste idea for infrared dim and small object augmentation. Moreover, they further proposed to copy the objects as well as their local background together so that retain as much local context of pasted objects as possible. In light of the infrared sensing mechanism is based on the temperature difference between the background and objects, the pasted objects may be incompatible with the novel backgrounds. To solve this problem, Wang et al. ~\cite{wang2022prior} first assumed that the boundary generation of dim and small objects follows the Gaussian blurring process and proposed to approximate this process during pasting, and then presented a gray correction method based on the gray distribution prior.  Note that the above augmentation strategies consider only a single type of scene (such as maritime object recognition). However, practical applications usually need to tackle multiple types of scenes. For example, drones may need to detect both aircraft objects and maritime objects simultaneously. In this case,  how to ensure the authenticity of the scene context (e.g., we cannot paste a ship into the sky) after augmentation becomes the key problem~\cite{yang2020small}, which existing methods have not yet considered due to the lack of infrared datasets. 

\subsubsection{Weighted Loss} The core idea of weighted loss is to assign more weight to tail classes when computing the training loss. Since most datasets in Table \ref{table_dataset} formulate dim and small object recognition as a binary segmentation problem, many methods have directly employed the soft IoU loss suited for imbalanced binary segmentation tasks. The soft IoU loss views the predicted probability as a soft label and then computes the intersection over union respect to the truth one-hot label. In particular, denoting  the predicted 2D probability map and 2D label map by $P$ and $Y$, respectively, the SoftIoU can be computed by 

\begin{equation}
 L=1-\frac{s+\sum P \odot Y}{s+\sum P+\sum Y-\sum P \odot Y}
 \label{eq_softiou}
\end{equation}

\noindent where $\odot$ represents handmard product.

Inspired by the success of Dice loss~\cite{milletari2016v} achieved in small medical object segmentation, Qi et al. ~\cite{qi2022ftc} used Dice Loss for infrared dim and small object recognition. Note that there is no essential difference between Dice Loss and IoU Loss. Both emphasize the overall similarity between the segmented mask and the truth object mask and pay less attention to how to ensure such overall similarity with pixel-level supervision. To this end, based on the fact that dim and small objects are more difficult to recognize than backgrounds, Chen et al.~\cite{chen2022multi} introduced the idea of focusing on hard samples of Focal Loss~\cite{lin2017focal} for infrared perception and applied it together with the IoU loss. However, the IoU loss and Foal loss are fused through a simple addition, which may lead to conflicts between the overall and pixel-level supervision. In general, the Focal loss can be computed by 

\begin{equation}
L= -\left(1-p_t\right)^\gamma \log \left(p_t\right)
 \label{eq_focalloss}
\end{equation}
\noindent where $p_t$ is the prediction on sample $t$. Note that $-log(p_t)$ is the standard cross-entropy loss when considering a one-hot label. Thus, the hard-sample focusing mechanism is achieved by viewing the samples with low predicted probabilities as hard samples and then assigning more weights to these samples to achieve a dynamic balance between foregrounds and backgrounds. 

Recently, Liu et al.~\cite{liu2024infrared} argued that the widely used IoU and
Dice losses lack sensitivity to the scales and locations of
targets, and developed a novel Scale and Location Sensitive
(SLS) loss for infrared small object recognition. The SLS loss achieved scale sensitivity by adding a scale-depended weight to the commonly used IoU loss, and achieved local sensitivity by forcing the predicted target centers to follow with the ground truth centers with a penalty term over center prediction.


\begin{table}
\renewcommand\arraystretch{1.4}
\caption{Focus and methods of infrared small object recognition, where the blank entries of columns 2-6  mean the corresponding methods did not give many words or not highlight it as their main contributions. }
\scriptsize
\centering
\hspace*{-2.5cm}
\begin{tabular}{llllll}
\hline
\multirow{3}{*}{Methods} &\multicolumn{5}{c}{Focused Challenges and Proposed Stategies}\\
\cline{2-6}
   &\multicolumn{3}{c}{Modeling of Infrared Small Objects} &\multicolumn{2}{c}{Long-tailed Training}\\
\cline{2-6}
  &Fusion Refinement &Local Refinement. &Semantics Refinement  &Loss Function &Data Aug.\\
\hline
TCRNet~\cite{mcintosh2020infrared} & &Offline learned layers &  & &\\
Dai et al.~\cite{dai2021asymmetric}  & ACM & & & &\\
Huang et al.~\cite{huang2021infrared} & FAMCA & &LSPM  &BCE &\\

UIUNet~\cite{wu2022uiu} &IC-A &Eembedded UNet &  & &\\
ISNet~\cite{zhang2022isnet} &TOAA & &  &Dice &\\
DNANet~\cite{li2022dense} &Nested Fusion& CSAM & Pyramid &SoftIoU &\\
ALCLNet~\cite{yu2022pay} & SBAM & & &SoftIoU &\\
FTCNet~\cite{qi2022ftc} & &Transformer &Transformer  &Dice Loss &\\
Wang et al.~\cite{wang2022prior} & & &  & &Copy-Paste\\
Chen et al.~\cite{chen2022multi} & & &  &Focal +Multi-task &\\

IR-TransDet~\cite{lin2023ir} & & &ASPP+Transformer  &SoftIoU &\\
MTU-Net~\cite{wu2023mtu} & & &Transformer  &FocalIoU &Copy-Paste\\
AGPCNet~\cite{zhang2023attention}  & AFM & &AGCB  &SoftIoU &\\
LCAGNet~\cite{nian2023local}  &MCAF&MALC &  &SoftIoU &\\
Liu et al.~\cite{liu2023infrared} & & &Transformer  &SoftIoU &\\

FcNet ~\cite{guo2024fcnet}   &CFM &FEM & SFM &SoftIoU &\\
GANet~\cite{zhang2024global} &AAF& Transformer&  &Focal+IoU &Copy-Paste\\
Zhang et al.~\cite{zhang2024infrared} &Adjacent Fusion & ICA& CBAM & &\\
NLFPNet~\cite{xi2024non} &AAFM& & NLPPM & &\\
AFENet~\cite{wang2024afe} &EDIG& & CFAR &SoftIoU &\\
EFLNet~\cite{yang2024eflnet} & & &  &Threshold focal &\\
HMNet~\cite{quan2024hmnet} &  & SAE & HAA &SoftIoU &\\
Chen et al.~\cite{chen2024feature}   &OCDF &Dual Encoder &  &BCE+Dice &\\
DMFNet~\cite{guo2024dmfnet}  &TCFM &Dual Encoder & REAM & SoftIoU &\\
SCTransNet~\cite{yuan2024sctransnet} &CCA & SCTB&  &BCE &\\
HAMANet~\cite{zhong2024hierarchical} &CFM &SiAM/PiAM & CGMLP &BCE+Dice &\\
MDAFNet~\cite{li2024moderately} & &MDAF &  &SoftIoU &\\
GSTUNet~\cite{zhu2024towards} & &Swin Transformer+GCL &ASPP  &Dice +BCE  &\\
CMNet~\cite{li2024cross} &  & MRCB & MRCB & &\\
MSHNet~\cite{liu2024infrared} &  &  &  &SLS &\\

\hline


\hline

\end{tabular}
\label{table_allmethods}
\end{table}

\subsection{Summary and the Future}
In summary, existing datasets for infrared dim and small object recognition are significantly smaller than datasets for visible light perception. As a result, existing methods usually train the feature model and the recognition model end-to-end over a specific small dataset, making them can hardly be generalized to diverse scenes or tasks. The efforts cared about training mainly focused on solving the long-tailed problem caused by the dim and small properties of interested objects. The data augmentation and weighted loss then received great attention. However, existing data augmentation methods designed for a specific dataset have not yet considered the context consistency constraints when confronted with diverse infrared scenes, and how to achieve a trade-off between overall and pixel-level supervision in weighted loss is still an open issue, leaving much room for future improvement. In addition, according to the development experience of visible light perception, how to achieve large-scale unsupervised pre-training that can be generalized to diverse infrared scenes may be the next potential hot spot. 

\section{State-of-the-arts}
Note that feature modeling and model training mentioned in Section \ref{sec_featuremodeling} and Section \ref{sec_modeltraining} are the two fundamental problems for deep learning-based recognition. Thus, many methods have jointly employed the strategies mentioned in these two sections. To this end, we have summarized the methods mentioned in these two Sections in Table \ref{table_allmethods}.  In addition, we conduct a quantitative comparison between representative SOTA methods to make readers quickly get the state-of-the-art of infrared small object segmentation. Since the experimental settings, including dataset, training settings, and evaluation metrics, are different in existing methods, most methods usually reproduce their counterparts using the same experimental setting for a fair comparison.  To this end, we chose one of them that has evaluated many methods over the most widely used datasets and open-sourced the code to simplify the fair comparison between SOTA methods. In particular, we excerpt the performance of 6 well-known methods, including ACM, ALCNet, MTU-Net, DNANet, UIU-Net, AGPCNet, reproduced by SCTrasNet. Moreover, we further reproduce 3 recent methods (including MSHNet, MRF3Net,HCF-Net) that have made their code publicly available, by ourselves using the same settings as SCTransNet.  The reproduction is conducted with a computer equipped with a GPU of NVIDIA RTX 4090 (24G), a CPU with Intel I9-14900k, and 64G RAM. We chose the most frequently used metrics (including normalized intersection over union (nIoU), Probability of Detection ($P_d$), and False alarm rate ($F_a$)) for comparisons. We also compared the efficiency of different methods using the number of parameters and floating
point of operations (FLOPs). The final results of the comparison are presented in Table \ref{table_sota}.  
\subsection{Commonly used metrics}
\subsubsection{Intersection over Union (IoU)}
The IoU is commonly used in segmentation tasks, denoting $A_i$ and $A_u$ the number of samples of intersection and union of prediction and ground truth, the IoU can be computed by
\begin{equation}
 \mathrm{IoU}=\frac{A_i}{A_u}=\frac{\sum_{i=1}^N \mathrm{TP}[i]}{\sum_{i=1}^N(T[i]+P[i]-\mathrm{TP}[i])}
 \label{eq_iou}
\end{equation}

\noindent where $TP[·]$ denotes the number of true positive pixels, $T[·]$ and $P[·]$ represent the number of ground truth and predicted positive pixels, respectively. With the same denotation, the normalized IoU is computed by 
\begin{equation}
\mathrm{nIoU}=\frac{1}{N} \sum_{i=1}^N \frac{\mathrm{TP}[i]}{T[i]+P[i]-\mathrm{TP}[i]}
 \label{eq_niou}
\end{equation}
\subsubsection{Probability of Detection($P_d$)}
Different from IoU, which computes at pixel-level, $P_d$ is a target-level evaluation metric and it measures the ratio of correctly predicted targets $N_{correct}$ over all targets $N_{all}$ as follows
\begin{equation}
P_d=\frac{N_{correct}}{N_{all}}
 \label{pd}
\end{equation}
\noindent The ``correctly predicted" is usually determined by the centroid deviation (e.g., less than 3 pixels~\cite{li2022dense}) between the prediction and the ground truth. 

\subsubsection{False alarm rate ($F_a$)}
$F_a$ measures the ratio
of falsely predicted pixels $P_{false}$ over all image pixels $P_{all}$ as follows
\begin{equation}
F_a=\frac{P_{false}}{P_{all}}
 \label{fa}
\end{equation}

\begin{table}
\renewcommand\arraystretch{1.4}
\caption{ The quantitative (nIoU ($10^{-2}$), $P_d (10^{-2})$, and $F_a (10^{-6})$) comparison of state-of-the-arts on three datasets. The Ref. and Aug. represent refinement and augmentation, respectively.}
\scriptsize
\centering
\hspace*{-2.5cm}
\begin{tabular}{l | l | l |  lll |  lll | lll}
\hline
\multirow{2}{*}{Methods} & \multirow{2}{*}{Paras./FLOPs} &\multirow{2}{*}{Strategies}  &\multicolumn{3}{c}{NUDT-SIRST}  &\multicolumn{3}{c}{NUAA-SIRST} &\multicolumn{3}{c}{IRSTD-1K}\\
& & &nIoU &$P_d$ &$F_a$  &nIoU &$P_d$ &$F_a$  &nIoU &$P_d$ &$F_a$\\
\hline
ACM (WACV/2021) ~\cite{dai2021asymmetric} & 0.40M/0.40G &Fusion Ref. &64.40 &93.12 &55.22  &69.18 &91.63 &15.23  &57.03 &93.27 &65.28\\
\hline
ALCNet(TGRS/2021)~\cite{dai2021attentional} & 0.43M/0.38G &Fusion Ref. &&&&&&&&&\\
 & &Local Ref.&67.20 &94.18 &34.61  &71.05 &94.30 &36.15 &57.14 &92.98 &58.80\\
\hline

 &  &Fusion Ref. &&&&&&&&&\\
DNANet(TIP/2022)~\cite{li2022dense} &4.70M/14.26G &Local Ref. & 88.58 &98.83 &9.00  &79.20 &95.82 &8.78 &66.38 &90.91 &12.24\\
& &Semantic Ref.  &&&&&&&&&\\
\hline

 & &Fusion Ref. &&&&&&&&&\\
 UIU-Net (TIP/2022)~\cite{wu2022uiu}&50.54M/54.43G&Local Ref. &93.89 &98.31 &7.79  &79.99 &95.82 &14.13 &66.66 &93.98 &22.07\\

\hline
 & &Semantic Ref. &&&&&&&&&\\
MTU-Net (TGRS/2023)~\cite{wu2023mtu}&  8.22M/6.20G & Data Aug. &77.54 &93.97 &46.95  &78.27 &93.54 &22.36 &63.24 &93.27 &36.80\\
& &Curated Loss &&&&&&&&&\\
\hline

&&Fusion Ref. &&&&&&&&&\\
AGPCNet(TAES/2023)~\cite{zhang2023attention}& 0.48M/0.41G & Semantic Ref.&90.64 &97.20 &10.02  &76.60 &96.48 &14.99 &65.23 &92.83 &13.12\\
&&(Self-Atten.) &&&&&&&&&\\
\hline

&&Fusion Ref. &&&&&&&&&\\
SCTransNet (TGRS/2024)\cite{yuan2024sctransnet}&11.19/20.24G& Local Ref. &94.38 &98.62 &4.29  &81.08 &96.95 &13.92 &68.15 &93.27 &10.74\\
&&(Transformer) &&&&&&&&&\\

\hline
& &Fusion Ref. &&&&&&&&&\\
MRF3Net (TGRS/2024)\cite{zhang2024mrf} & 0.54M/8.30G &Local Ref.  &95.12 &98.97 &3.29  &80.12 &96.65 &10.26 &68.14 &96.16 &12.56\\ 
\hline
& &Fusion Ref. &&&&&&&&&\\
HCF-Net (arXiv/2024)\cite{xu2024hcf}&15.29M/23.29G & Local Ref. &93.11 &98.21 &7.07  &79.67 &95.95 &11.32 &68.05 &95.55 &11.13\\
& &Semantic Ref. &&&&&&&&&\\
\hline
MSHNet (CVPR/2024)\cite{liu2024infrared}& 4.07M/6.08G&Curated Loss &91.92 &98.61 &8.99  &79.18 &96.03 &14.92 &67.92 &94.17 &13.74\\
& &Multi-scale Head &&&&&&&&&\\


\hline

\hline

\hline

\end{tabular}
\label{table_sota}
\end{table}

\subsection{Analysis} 
Based on these evaluation metrics, an excellent infrared small object segmentation model should produce a high value on both $nIoU$ and $P_d$, while producing a $F_a$ as low as possible.  Based on this insight, several important observations can be obtained from Table \ref{table_allmethods} and Table \ref{table_sota} as follows. 
\begin{enumerate}
\item Without loss of generality,  the performances achieved on the IRSTD-1K dataset are lower than those achieved on the NUDT-SIRST and the NUAA-SIRST. For example, all the nIoUs achieved on IRSTD-1K are less than 70\%, while the highest nIoU achieved on NUDT-SIRST and NUAA-SIRST reach 95.12\% and 81.08\%, respectively. One main reason is that IRSTD-1K contains more types of scenes and categories, resulting in a high demand for the model's ability on generalization.  

\item From the perspective of accuracy, the performances achieved by MRF3Net, SCTransNet, and HCF-Net can be used to represent the SOTA of infrared small object segmentation.  For the SOTA of NUDT-SIRST, the nIoU and $P_d$, are greater than 93\% and 98.5\%, respectively,  and $F_a$ is less than $10^{-6}$. For the SOTA of NUAA-SIRST, the nIoU and $P_d$, are greater than 79.5\% and 95.9\%, respectively,  and $F_a$ is less than $15 \times 10^{-6}$.  For the SOTA of IRSTD-1K, the nIoU and $P_d$, are greater than 68\% and 93\%, respectively,  and $F_a$ is less than $13 \times 10^{-6}$. 

\item Most methods published after 2021 employ at least two strategies mentioned above to improve feature modeling, which also suffers the disadvantage of being computationally expensive. For example, the UIU-Net, which achieved the SOTA performance before 2024, produced the largest number of FLOPs (54.43G). Although the FLOPs is greatly reduced by later SOTA methods such as MRF3Net (8.30G), the computational costs of these methods are still larger than earlier baseline (e.g., ACM) by dozens of times.

\item The MTU-Net, which employs simultaneously semantic refinement, data augmentation, and curated loss, lags behind its counterparts that use multiple refinement strategies. An explanation for this is that the MTUNet is specially designed for space-based infrared tiny ship detection, especially the data augmentation strategy. 

\item Most recent methods achieved a $P_d$ value larger than 95\% and a $F_a$ value less than $15\times 10^{-6}$ in small-scale datasets containing scenes of limited types, such as NUDT-SIRST and NUAA-SIRST; it can be expected that the performance gaps between methods will be negligible in a short time. According to the development trajectory of semantic segmentation on visible light images, future attention on infrared small object segmentation will mainly focus on developing models that can be generalized to any types of scene.

\end{enumerate}

\color{black}
\section{Model Transfer} 
In general, a visual perception system usually needs to tackle an open task in practical applications, which means the objects that need to be recognized will change with the change of scenes. Therefore, how to transfer directly an existing perception model to a novel but similar scene is always one of the key problems in visual perception~\cite{chen2023transfer,yao2023survey,yang2024domain}. This section will introduce transfer learning methods in infrared small object segmentation, including few-shot learning, unsupervised domain adaptation, and large foundation models-based transfer.  


\subsection{Few-shot Learning}
Few-shot learning aims to mimic humans’ ability to recognize object classes once they have seen only one or a few instances of each class. For example, Chen et al. applied the famous meta-learning framework MAML~\cite{finn2017model} to few-shot infrared perception. This method used ground infrared classes that can be easily obtained as the base classes and then simulated the few-shot recognition process with samples of the base classes. Then, the learned few-shot recognition experience was used to guide the fine-tuning of novel (aerial) class recognition.  To improve the model's ability to learn common knowledge of classes from few-shot samples of them, Maliha Arif and Abhijit Mahalanobis employed class-agnostic filtering to extract a common manifold representation for diverse scenes~\cite{arif2021few}. In particular, the class-agnostic filter is obtained by first performing feature clustering and then solving the maximum eigenvector of the clustering results.  The computed maximum eigenvector was then applied to extract class-agnostic local salient features, which greatly reduced the difficulty of subsequent few-shot learning. The base classes and novel classes used in the above two methods are taken from the same infrared dataset. Miao et al.~\cite{miao2023few} proposed to use large-scale visible light scenes to learn a few-shot recognition model and then fine-tune the model for infrared object recognition. 
\begin{table}
\renewcommand\arraystretch{1.5}
\caption{ FSS means few-shot samples, AT.FA means adversarial training-based feature alignment, AT.IT means adversarial training-based image translation, ST means self-training }
\scriptsize
\centering
\hspace*{-3.5cm}
\begin{tabular}{lll}
\hline
Methods &Ideas and Innovations. &Transfer Path \\
\hline
Few-shot learning: & &\\
\hline

Miao et al.~\cite{miao2023few}& MAML (Pre-train on the base classes and then finetune head network with FSS) & Infrared-to-Infrared\\
Mahila ~\cite{arif2021few} &Similar to the above methods, but employ an analytically designed low-level convolution & Infrared-to-Infrared\\
Tai et al.~\cite{tai2023mine} &Learn to extract class prototypes from FSS on base classes & Infrared-to-Infrared\\

\hline
Domain Adaptation:& &\\
\hline
Lu et al. ~\cite{gan2023unsupervised}& apply AT.FA between $S$ and $T$  & RGB-to-Thermal\\
Zhang et al.~\cite{zhang2023two} & apply AT.IT (source to target), and then performed a multi-model based training   &RGB-to-Infrared \\
Kim et al. ~\cite{kim2021ms} & apply mutual AT.IT between $S$ and $T$ & Thermal-to-Thermal \\
Chen et al.~\cite{chen2022light} & apply ST: generating pseudo labels on $T$ with models learned from $S$ &RGB-to-Thermal\\
Akkaya et al.~\cite{akkaya2021self} & apply AT.FA and ST jointly &RGB-to-Thermal\\
Chi et al.~\cite{chi2024semantic} & perform supervised training on the S and T alternatively with a shared feature network.  &Infrared-to-Infrared\\


\hline

\end{tabular}
\label{table_modeltransfer}
\end{table}
Note that the above methods pay less attention to the recognition accuracy of base classes, which can easily lead to a catastrophic forgetting problem of base classes. However, practical applications usually need to retain the performance of base classes after introducing novel classes. To this end, Tai el al.~\cite{tai2023mine} presented an incremental few-shot learning framework for infrared perception. To achieve incremental recognition of classes ( base class + all novel classes), based on the class prototype representation introduced in ~\cite{snell2017prototypical}, this framework formulated the few-shot learning problem as a feature matching between query features and the ``class prototypes” extracted from the few-shot supported samples. Moreover, to improve the accuracy of prototype representation, this method first extracted a high-dimensional prototype and then distilled a prototype in a lower dimension that consists of key channels that affect the few-shot recognition the most.   

Although the above few-shot learning methods have achieved great success for infrared perception, one of the main drawbacks is they all assume that there are a large number of base class samples to support few-shot learning, which usually cannot be ensured in practical applications.

\subsection{Domain Adaptation}
Domain Adaptation refers to a task that learns a recognition model from a source domain containing rich annotations and applies it to unlabeled target domain recognition.  Due to the lack of available large-scale infrared datasets, some methods instead resort to virtual source domains for infrared domain adaptation. However, the generated virtual data are too ideal compared to real infrared data. Therefore, domain adaptation methods for infrared perception usually utilize the visible domain as the source domain. For example, Lu et al.~\cite{gan2023unsupervised} proposed an unsupervised domain adaptation method that transfers knowledge from the visible domain to the thermal infrared domain. This method adopted a shared backbone with domain-dependent attention to extracting features from two domains separately and then aligned the features of the two domains by performing adversarial training. Zhang et al.~\cite{zhang2023two} instead applied adversarial training to translate the infrared images into visible light images. The translated visible light images are then concatenated with real visible light images for training. To avoid the long-span translation from visible light to infrared images, Kim et al.~\cite{kim2021ms} proposed to perform mutual translation between daytime infrared images and nighttime infrared images for nighttime traffic scene recognition.

Although adversarial training can achieve multi-level (e.g., input, feature, and output) alignment of two domains, it is notorious for its sensitivity to hyperparameters. To avoid the use of adversarial training, another well-known domain adaptation strategy named self-training applies first the recognition model learned from the source domain to generate pseudo labels on the target domain and then performs training over the target domain with these pseudo labels. For example, based on the observation that semantic contours are less affected by domain distribution, Chen et al.~\cite{chen2022light} proposed to use the semantic contours to generate pseudo-labels for the target infrared domain. 

Although the idea of self-training is simple and intuitive, the hard samples cannot be effectively trained since prediction over them always has low confidence. To alleviate this problem, Akkaya et al.~\cite{akkaya2021self} used adversarial training and self-training jointly for domain adaptation from visible light to thermal infrared. In particular, to improve the accuracy of pseudo-labels generated by visible recognition models, this method applied adversarial training to achieve feature alignment between the visible domain and infrared domain.

\begin{figure}[!h]
\centering
\includegraphics[width=\textwidth]{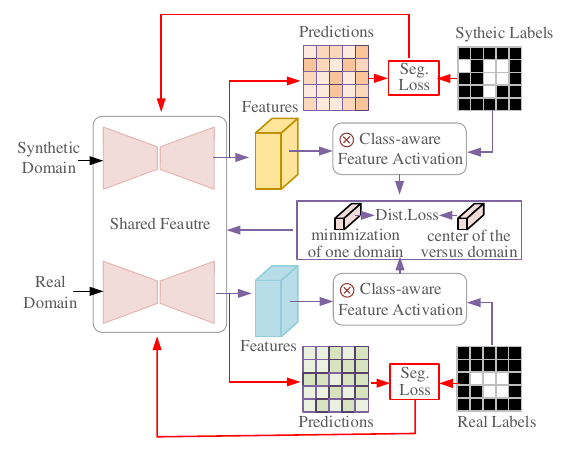}
\caption{The core idea of SDAISTD, which presents one of the first domain adaptation methods for infrared small object recognition.}
\label{fig_dasdai}
\end{figure}

It can be seen from Table \ref{table_modeltransfer} that most domain adaptation methods for infrared perception focused on short-distance thermal infrared scenes. Recently, Chi et al.~\cite{chi2024semantic} proposed a pioneer domain adaptation framework coined SDAISTD for long-distance infrared perception. Note that SDAISTD performs domain adaptation in a supervised manner, i.e., the target domain is also labeled, but not in the mainstream unsupervised manner (Fig. \ref{fig_dasdai}).  In particular, the SDAISTD performed supervised training on the source domain and target domain alternatively with a shared feature extraction network. To minimize the feature distance between the different domains during this training, the SDAISTD defined an intra-class distance function by computing the Euclidean distance between the minimized feature of one domain and the feature center of its opposite domain. 
\subsection{Transfer from Large Foundation Models}
With the recent great success achieved by multi-modal large foundation models such as CLIP, and SAM, directly transferring the universal knowledge of foundation models for vision tasks has received increasing attention. However, it is difficult to transfer knowledge of existing large models learned from massive visible-light data to the infrared domain due to the domain gap. A pioneer work coined IRSAM ~\cite{zhang2024irsam} proposed to address the domain gap problem by introducing two specialized feature enhancement modules to the SAM'S encoder-decoder architecture before finetuning it for infrared small object segmentation. In particular, a wavelet transform-based Perona-Malik diffusion was first used to enhance the encoder's ability to suppress noise while preserving smooth structural features, and a multi-granularity-based fusion was used to enhance the decoder's ability to preserve both global context and local details for the final recognition. While large foundation models have not yet received great attention in infrared small object segmentation, they have been widely used in research fields such as medical small object segmentation and small defect detection that share similar challenges to the infrared small object segmentation. Next, we will introduce some representative large foundation model based methods in these related research fields, which aims to provide possible inspirations for this important future direction.  

Without loss of generality, fine-tuning is the most classical method to employ large foundation models for professional scenes that are different from data used for training these large models.  This line of method utilizes the foundation model as a pre-trained feature extractor, then fine-tuning it with the professional data. For example, the IRSAM mentioned above belongs exactly to this class. However, performing full fine-tuning, i.e., fine-tuning all the parameters of the foundation model, is computationally expensive, and moreover, data-hungry. For example, the MedSAM~\cite{ma2024segment} creates a large and varied dataset to fine-tune the SAM for medical image segmentation. While obtaining a large amount of professional data is usually difficult, the parameter-efficient fine-tuning received more attention. The parameter-efficient fine-tuning~\cite{wu2023medical,huang2024adapting} usually keeps the parameters of the foundation models frozen, and then incorporates several learned adapters to tailor the foundation models for professional scenes. For example, to adapt SAM to medical images, the Med-SA~\cite{wu2023medical} positioned an MLP-based bottleneck adapter at the self-attention path and residual connection path of the ViT block in SAM, respectively. Such adaptation was then extended to 3D medical images by transposing the spatial dimension of 3D medical input embedding to the depth dimension. Considering SAM is a prompt-based interactive segmentation model, instead of incorporating adapters into the basic feature block of SAM, AutoSAM~\cite{shaharabany2023autosam} proposed to learn an image-based prompt encoder to adapt frozen SAM for medical images. 

Considering that most foundation models offer aligned visual-text representation, aligning the novel visual patterns of professional scenes to well-designed text prompts to achieve zero-shot visual recognition has also been explored in related fields. For example, to achieve zero-shot anomaly segmentation, the WinCLIP~\cite{jeong2023winclip} first applies visual feature extraction at multi-scale windows using CLIP and then guides the feature recognition with a compositional prompt ensemble composed from a pre-defined list of state words and varied text templates. Since existing foundation models are trained on natural image-text datasets that differ significantly from professional data, models like WinCLIP which introduce only curated hand-crafted prompts typically exhibit limited performance. To this end, adapting foundation models on auxiliary datasets is gaining popularity for zero-shot learning tasks. Note that the auxiliary datasets can not contain categories awaiting recognition with zero-shot annotations.  For example, Cao et al. ~\cite{cao2025adaclip} defined a hybrid learnable prompt that consists of an image-dependent dynamic prompt and an image-independent static prompt. The hybrid prompt is then concatenated with the visual and text tokens of the frozen foundation model and trained on the auxiliary dataset for better visual-text alignment for industrial and medical scenes.  

Since the exploration of large foundation models in the area of infrared small object segmentation has not received great attention, this subsection instead introduces the most representative foundation model-based methods in related fields. For more works about the large foundation models as well as their applications to related visual analysis tasks,
we refer readers to several excellent reviews such as \cite{wu2024towards,ali2024review}.

\subsection{Summary and the Future}
It can be observed from the above methods that both few-shot learning and domain adaptation rely on a large-scale base dataset (e.g., the base classes for few-shot learning, and the source domain for domain adaptation) for model training. Thus, due to the lack of large-scale datasets, there are still very few model transfer methods that contribute to long-distance infrared perception. This paper instead reviews some related domain transfer methods on short-distance infrared perception. In other words, the model transfer for long-distance infrared perception is still an unexplored or less-explored area.  Based on the current trends in visible light sensing, developing a foundation model suitable for infrared perception and then achieving zero-shot transfer for infrared sensing with the foundation model will be a promising future direction.

\section{Discussion and Conclusion}
This paper provides a short while structural review of challenges and recent achievements in deep learning-based infrared dim object recognition.  We mainly focused on providing a clear analysis of existing challenges, and then clarifying both the motivations of existing methods and potential future
directions from the perspective of these proposed challenges. Based on this review, it can be concluded that existing methods mainly focus on challenge issues such as feature modeling, and long-tail learning. Innovative strategies such as multi-scale feature fusion, Transformer-based context modeling, copy-and-paste data augmentation, and Soft IoU-based weighted loss have made important progress on small-scale datasets designed for specific types of infrared scenes. 

In the context of the applications of long-distance infrared perception has become increasingly broad, existing methods suffer from drawbacks such as not fully differentiating the modeling of signal from noise, having a computationally expensive multi-scale feature enhancement, not fully considering the scene context restricts for data augmentation in diverse scenarios and the trade-off between overall and pixel-level supervision when designing weighted loss. In addition, based on common experience in natural language processing and visible light perception, using large-scale pre-training to learn common knowledge is the key to supporting the generalization of the perception model to different scenarios. Given that short-distance infrared images and visible light images share the same scene structure and semantic context dependency, the above experience has been extended to the field of short-distance infrared perception. On the contrary, model pre-training for long-distance infrared scenes has not received enough attention. Thus, future works can be conducted by solving these mentioned drawbacks and paying more attention to the less-explored pre-training and transferring problems.

\section*{Acknowledgement}
This work was supported in part by the National Natural Science Foundation of China under Grant 62103137, U2013203, 62373140, in part by the Natural Science Fund of Hunan Province (2022JJ40100,2021JJ10024), the Key Research and Development Project of Science and the Technology Plan of Hunan Province under Grant 2022GK2014, National Key R\&D Program(2023YFB4704503), and Leading Talents in Science and Technology Innovation of Hunan Province(2023RC1040), Professor Ajmal Mian is the recipient of an Australian Research Council Future Fellowship Award (project number FT210100268) funded by the Australian Government.

\bibliography{mybibfile_nodoi}

\end{document}